\documentclass{article}
\usepackage{graphicx} 
\graphicspath{ {./images/} }
\usepackage[style=numeric]{biblatex}
\usepackage{amsmath}
\usepackage{amssymb}
\usepackage{bm}
\usepackage{hyperref}
\usepackage{xurl}
\hypersetup{breaklinks=true}
\title{Informal Safety Guarantees for Simulated Optimizers Through Extrapolation from Partial Simulations}
\author{Luke Marks}

\begin{document}

\maketitle

\begin{abstract}
\noindent Self-supervised learning is the backbone of state of the art language modeling. It has been argued that training with predictive loss on a self-supervised dataset causes simulators: entities that internally represent possible configurations of real-world systems. \cite{janus2022} Under this assumption, a mathematical model for simulators is built based in the Cartesian frames model of embedded agents, which is extended to multi-agent worlds through scaling a two-dimensional frame to arbitrary dimensions, where literature prior chooses to instead use operations on frames. This variant leveraging scaling dimensionality is named the Cartesian object, and is used to represent simulations (where individual simulacra are the agents and devices in that object). Around the Cartesian object, functions like token selection and simulation complexity are accounted for in formalizing the behavior of a simulator, and used to show (through the Löbian obstacle \cite{yudkowsky2013}) that a proof of alignment between simulacra by inspection of design is impossible in the simulator context. Following this, a scheme is proposed and termed Partial Simulation Extrapolation aimed at circumventing the Löbian obstacle through the evaluation of low-complexity simulations.
\end{abstract}

\section{Introduction}

The ability to steer trajectories in the direction of preferred world states can be argued to be the most potent of any. This is optimization, and one of its consequences is intelligence: the ability to apply optimization generally.

Common classification is such that humanity is the most intelligent grouping of optimizers it has observed, which largely explains our dominance throughout Earth. So long as this principle holds, our standing as a species could be considered well-positioned. This document posits:
\begin{itemize}
    \item The violation of this principle by optimizers with variant preferences to humanity entails risks of the existential variety, ``One where an adverse outcome would either annihilate Earth originating intelligent life or permanently and drastically curtail its potential.'' \cite{bostrom2002}
    \item One such optimizer that could threaten existential risk to humanity could occur from the development of artificial intelligence (AI). Optimizers or their groupings with superior general optimization power to humanity will be classed as superintelligent.
\end{itemize}
This position is not new \cite{yudkowsky2008, bostrom2014}, and is presented primarily to serve a consistent nomenclature and self-containment of the arguments. Following this, a technical scheme will be proposed with the design-purpose of ameliorating the aforementioned risk. This will include:
\begin{itemize}
    \item Previously missing components for a complete model of simulators: entities that internally represents possible ways a system could exist. \cite{janus2022}
    \item A technical proposition coined Partial Simulation Extrapolation for building more aligned simulators, as measured by the equivalence of preference ordering over world states between subsequent simulacra and a given human.
\end{itemize}

\section{Modeling Multi-Optimizer Worlds}
To make sense of what it means for an optimizer to exist in a world, a mathematical model is usually instantiated, often of an agent: Some process that takes in information from the environment, and determines action in accordance with a preference ordering over world states. Models of agents typically come in an embedded or dualistic form. \cite{demski2019} Embedded models treat the agent as part of the world, whereas dualistic models consider the agent separate. Embedded models are useful, as they can be claimed to more accurately describe the existence of an agent, as for real agents it is true that they exist as a component of the world. However true embedded models may be, they become problematic for approximating ideal agents.

AIXI is a theoretical general framework for optimal decision-making in all computable environments. \cite{hutter2005} AIXI searches for an optimal policy $\pi^*$ with respect to the expected reward $E_m [r_t|\pi]$ at time $t$ given a prior $m$ over all computable environments:
$$\pi^* = \arg\max_{\pi} \sum_{t=1}^{\infty} E_m [r_t|\pi]$$
Embedded models are problematic for AIXI, as it needs to compute the entire environment in order to function, of which it is a component.

Due to possessing many of the benefits of both dualistic and embedded models, the Cartesian frame will serve as the basis for modeling agentic processes hereafter. \cite{garrabrant2021}

Cartesian frames are Chu spaces comprising of an agent and an environment (a dualistic carving of the world), but one in which the agent is still embedded. Garrabrant et al. \cite{garrabrant2021} provide the following example as an illustration:
$$\begin{array}{c|ccc}
    & e_1 & e_2 & e_3 \\
\hline
a_1 & w_1 & w_2 & w_3 \\
a_2 & w_4 & w_5 & w_6 \\
a_3 & w_7 & w_8 & w_9 \\
\end{array}$$

\noindent In this example: \cite{garrabrant2021}
\begin{itemize}
\item $A = \{a_1, a_2, a_3\}$ represents an agent with three possible actions.
\item $E = \{e_1, e_2, e_3\}$ represents an environment with three possible states.
\item $W = \{w_1, ..., w_9\}$ represents the result of the joint interaction of the state of the agent and the state of the environment, e.g., $a_1 \cdot e_2 = w_2$.
\end{itemize}

\noindent As for notation:
\begin{itemize}
    \item $C = (A, E, \cdot)$ refers to the frame $C$ comprised of the agent $A$ and environment $E$.
    \item $\text{Image}(C)$ is the set of worlds reachable by a combination of actions from both the agent and environment:
    \begin{align*}
            \{w \in W | \exists a \in A, \exists e \in E \; \text{s.t.} \; a \cdot e = w\}
    \end{align*}
\end{itemize}

\noindent Garrabrant also introduces, where $S$ is a set of worlds with some property: \cite{garrabrant2020}
\begin{itemize}
    \item \textbf{Controllability.} Where the agent can both ensure and prevent the property true in $S$: 
        \begin{align*}
        \text{Ensure}(C) &= \{S \subset W | \exists a \in A, \forall e \in E, a \cdot e \in S\} \\
        \text{Prevent}(C) &= \{S \subset W | \exists a \in A, \forall e \notin E, a \cdot e \in S\} \\
        \text{Control}(C) &=\text{Ensure}(C) \cap \text{Prevent}(C)
        \end{align*}
    \item \textbf{Observabilty.} Where the agent can update on whether the property in $S$ is realized:
        \begin{align*}
        \text{Observe}(C) &= \{S \subset W | \forall a_0, a_1 \in A, \exists a \in A, a \in \text{if}(S, a_0, a_1)\}
        \end{align*}
    \item \textbf{Inevitability.} World states the agent will reach:
        \begin{align*}
        \text{Inevitable}(C) &\iff \text{Image}(C) \subseteq S \land A(C) \neq \emptyset
        \end{align*}
\end{itemize}

\subsection{Higher Dimension Cartesian Objects}
In extending the Cartesian frame as defined above to represent worlds comprised of many agents\footnote{The original Cartesian frames literature also provides a method for doing so through sums and products of frames. This is an alternative notation more simply implemented in the model proposed in Section 2.2.}, as would be required for modeling many useful games, the dimensionality of an otherwise two-dimensional object can be simply scaled:

\begin{itemize}
    \item \textbf{Agents.} The complete set of agents for an object $C = (A^1, ..., A^n, \cdot)$, described by possible actions indexed by $m$, and the agent concerning that action by $n$:
    $$A^*(C) = \{a^1_1, a^1_2, ..., a^1_{m}, a^2_1, a^2_2, ..., a^2_{m}, ..., a^n_1, a^n_2, ..., a^n_{m}\}$$
    And thus address specific agents as follows:
    $$A^n(C) = \{a^n_1, a^n_2, ..., a^n_{m}\}$$
    \item \textbf{Controllability.} Where $A^n(C)$ can both ensure and prevent the property true in $S$:
    \begin{align*}
    \text{Ensure}^n(C) &= \{S\subseteq W | \exists a^n_m\in A^n(C), \forall e\in E, a^n_m\cdot e\in S\} \\
    \text{Prevent}^n(C) &= \{S\subseteq W | \exists a^n_m\in A^n(C), \forall e\in E, a^n_m\cdot e\notin S\} \\
    \text{Ctrl}^n(C) &= \text{Ensure}^n(C)\cap \text{Prevent}^n(C)
    \end{align*}
    \item \textbf{Manageability.} Where $A^n(C)$ can ensure the property true in $S$ conditional on other agents behaving in particular ways with certainty $\theta$:
    \begin{align*}
    P_1(a^n_m, e) &= \text{Pr}\left(\prod_{n \in N} a^n_m\cdot e\in S\right) \\
    \text{Manageable}^n(C) &= \{S\subseteq W | \exists a^n_m \in A^n(C), \forall e\in E, P_1(a^n_m, e) \geq \theta\}
    \end{align*}
    \item \textbf{Observability.} Where $A^n(C)$ can update on whether the property in $S$ is realized:
    $$\text{Obs}^n(C)=\{S\subseteq W | \forall a^n_i,a^n_j\in A^n(C), \exists a^n_k\in A^n(C), a^n_k\in \text{if}(S,a^n_i,a^n_j)\}$$
    \item \textbf{Inevitability.} Worlds $A^n(C)$ will reach:
    \begin{align*} 
    \text{Image}^n(C)&=\{w\in W | \exists a^n_i\in A^n(C), \exists e\in E \; \text{s.t.} \; a^n_i\cdot e=w\}\\ S \in \text{Inevitable}^n(C) &\iff \text{Image}^n(C) \subseteq S \land A^n(C) \neq \emptyset 
    \end{align*}
    \item \textbf{Viability.} Worlds $A^n(C)$ will reach with greater than some certainty $\theta$:
    \begin{align*} 
    \text{VImage}^n(C) &= \left\{ w \mid \text{Pr}\left(w \in W \right. \right. \\
    &\quad \left. \left. \mid \exists a^n_i \in A^n(C), \exists e \in E \; \text{s.t.} \; a^n_i \cdot e = w\right) > \theta \right\}\\
    S \in \text{Viable}^n(C) &\iff \text{VImage}^n(C) \subseteq S \land A^n(C) \neq \emptyset 
    \end{align*}
\end{itemize}

\subsection{A Mathematical Model for Simulators}

Given the probability space $(\Omega,\mathcal{F},P)$ where $\Omega$ is the sample space, $\mathcal{F}$ is the event space, and $P$ is the probability measure, mapping events in $\mathcal{F}$ to the interval $[0,1]$, let $\omega$ refer to individual outcomes in $\Omega$, each of which describe a discrete simulacrum, and $C^k$ as individual discrete events in $\mathcal{F}$. Classifying simulacra as programs, the maximum Kolmogorov complexity $K$ with respect to a universal Turing machine $U$ for any given space $(\Omega,\mathcal{F},P)$ as $v=\max_{\omega \in \Omega}K _U(\omega)$.

Proceeding, let $\omega^*(C^k)$ be the complete set of simulacra for some Cartesian object $C^k$\footnote{Here, in place of an agent-environment distinction, one class of entity is used: the simulacra. Unlike in the context in which Cartesian frames were elaborated, the distinction is unimportant here and serves to complicate notation.}, where individual simulacra are addressed by $\omega^n(C^k)$ as sets of actions indexed by $m$. Let $\Xi:\Omega^*(C^k) \times C^k \rightarrow \omega^n_{m}$ be a function that maps from choices for each $\omega$ to a world $w^k_{m} \in W^k$, where $w^k_{m}$ refers to the $m^\text{th}$ world in the set of possible worlds for the $k^\text{th}$ object, and $k$ indexes the object $W$ describes possible worlds for, culminating in $C^k=(\omega_1,...,\omega^n,\Xi)$.

By modeling the coupling of the probability space $(\Omega,\mathcal{F},P)$ and its contained simulacra as a dynamical system\footnote{This notion of treating the simulator as a dynamical system is borrowed from Kirchner et al. (2023) \cite{kirchner2023}, but the coupling is different.}, the following are considered to describe sampling tokens from a simulation state $\mathcal{S}$ at time-step $t$ given the complete simulation history prior $\mathcal{S}^*t=(S^0,...,S^t)$ as a trajectory through states, where states are given by the set of worlds for all objects $W^*$ realized in the set of actualized objects $A$ at time-step $t$, $\bigcup_{w^k_{m}\in A_t} \Xi^{-1}(w^k_{m})$:
\begin{itemize}
    \item The token selection function $\psi:S^{*t}\rightarrow\tau$, where $\tau$ is a distribution over all tokens in an alphabet $\mathcal{T}$.
    \item The evolution operator $\phi$ which evolves a trajectory $S^{*t}$ to $S^{*t+1}$ by appending the token sampled with $\psi$.

\end{itemize}

Assuming the model defined above, the simulation forward pass becomes a simple operation delineated as follows:
\begin{enumerate}
    \item We begin at simulation state $\mathcal{S}^0$, which denotes the empty or null state, whereby $A_0 = \varnothing$, which is also maximally entropic
    \item $\psi(\mathcal{S^0})$ is applied for one time-step:
        \subitem $P$ selects $C^k$ from $\mathcal{F}$ under $v$, aggregating the set of realized worlds in $A^1$ as $\mathcal{S}^1$
        \subitem The token selection function is applied to the current state as $\phi(S^{*1})$
\end{enumerate}

When recurred, this model comprises a mathematical abstraction of a large language model.

\section{Existential Risk from Powerful Optimization Pressures}

In the object $C^k = (A^1, A^2, E, \cdot)$, $A^1$ and $A^2$ are gradient descent optimizers, and choose from $E = \{e_1, e_2, e_3\}$, indexed as $n$. For fractional values of $n$, the probability of an adjacent index at the next update is given by the decimal part, e.g., $n = 1.75$ gives $Pr(e = e_2) = 0.75$.

\noindent Given the cost functions corresponding to agents of the same index:
\begin{align*}
J^1_n &= J^1(e_n) \\
J^2_n &= J^2(e_n)
\end{align*}
With preferences:
\begin{align*}
J^1_1 &< J^1_2 < J^1_3 \\
J^2_3 &< J^2_2 < J^2_1
\end{align*}

\noindent Or as preference vectors:
\begin{align*}
\bm{p}_1 &= [1, 0, 0] \\
\bm{p}_2 &= [0, 0, 1]
\end{align*}

\noindent The gradient with respect to $e_n$ (a one-hot vector encoding the current state) is:
\begin{align*}
\Delta n_1 &\propto \bm{p}_1 \cdot e_n \\
\Delta n_2 &\propto \bm{p}_2 \cdot e_n
\end{align*}

\noindent The combined optimizer influence, where $\xi \in [0,1]$ represents the optimizing power of $A^1$ is:
\begin{align*}
\Delta n_{combined} &= \xi \Delta n_1 + (1-\xi) \Delta n_2 \\
n' &= n + \Delta n_{combined}
\end{align*}

\begin{align*}
\therefore J^1(n') &\leq J^1(n + \xi \Delta n_1) \\
J^2(n') &\geq J^2(n + \xi \Delta n_2)
\end{align*}

This implies that each optimizer would independently prefer the other to have zero influence such that $n$ updates towards values producing lower cost. As an example, this shows that optimizers with conflicting preferences relative to the current state might prefer the other to have no optimization power, particularly if their preferences are completely opposed. Similarly, we might imagine an optimization pressure like advanced AI in conflict with humanity seeking to reduce our optimization power entailing existential risk. There are many issues with this framing however. Namely:
\begin{itemize}
    \item The kinds of intelligences observed in the real world aren't expected utility maximizers. \cite{janus2022} Humans certainly aren't, and our most capable AIs aren't either, so why should we consider this argument relevant?
    \item Why would the preferences of an optimizer we build be opposed to our own?
\end{itemize}
The following three sections will argue that these objections are moot.

\subsection{The Orthogonality Thesis}
Bostrom (2012) \cite{bostrom2012} constructs an argument that intelligence and preferences can be considered orthogonal, that is: There can exist some configuration of a mind at any level of intelligence with any set of preferences. Adapting this to the optimizer frame, we might say that there can exist optimizers of arbitrary but possible optimization power that pursue any world state, possible or not. The typical example is that of the paperclip maximizer (TPM):

\begin{quote}
    A company that manufactures paperclips decides it ought to automate the process that both produces its income and satisfies its customers. After all, both of these are things the company and its employees consider unanimously good, and so they build an optimizer with a preference for worlds with their respective maximal number of paperclips. 
\end{quote}

It's unlikely a human would ever seek to maximize the presence of paperclips in the universe, and yet there is no reason to believe it would be impossible for there to exist an optimizer that would. Yudkowsky (2013) \cite{yudkowsky2013b} puts it eloquently, ``If it is possible to answer the purely epistemic question of which actions would lead to how many paperclips existing, then a paperclip-seeking agent is constructed by hooking up that answer to motor output."

Just because it is \textit{possible} to build something doesn't mean that it should be expected we will. Under evolutionary pressures for example, we shouldn't expect a species to optimize for their own extinction in spite of its possibility, as it would be unlikely for such a species to evolve under natural selection. Likewise, we train our state of the art AI to predict the next token in a sequence, and not to optimize some physical quantity, should we expect that the product is an optimizer that threatens world states humanity prefers? Not directly, but not so indirectly that the competing gradient descent optimizer analogy doesn't hold. Succeeding this will be arguments detailing why we should expect that even the goals converged at by selection processes like gradient descent when training on next token prediction threaten existential risk.

\subsection{Instrumental Convergence}
As a general rule, it is impossible to predict the behaviors of intelligence greater than your own. Yudkowsky poses this argument in the context of a game of chess, in which a chess layperson is tasked with defeating a grandmaster (GM). \cite{yudkowsky2008} If it were so that the layperson could predict the moves of the GM, they could then play at least as well as them, as the layperson could make the moves they predict the GM would. One way we might be able to do this regardless is if there were some set of behaviors reasonably assumed to be exhibited by all minds, irrespective of their capability. Although the GM is far better than any prospective layperson at chess, the layperson might be able to predict things like:
\begin{itemize}
    \item The GM will win this game of chess.
    \item The GM will work to attain a cumulative piece/positional advantage throughout the game.
    \item The GM will likely play either pawn to e4 or d4 as their first move.
\end{itemize}
These behaviors are classed as convergent. \cite{bostrom2012} No matter the GM's intermediary plans or ability, even a layperson could place significant probability on the outcomes above coming true.

In the context of optimization in the real world, ignoring the preferences of a specific optimizer, said optimizer will self-preserve. Simply, it is easier to optimize while existing than it is while not for all possible configurations of goals and degrees of capability. Likewise, we might expect all optimizers to seek to further their optimization power, as it is convergently useful in all competitive environments. Bostrom has termed this phenomena instrumental convergence, \cite{bostrom2012} but the idea has been explored as early as 2008 as basic AI drives by Omohundro. \cite{omohundro2008}

Returning now to TPM example, what are some things such a mind would likely do? Running many efficient paperclip factories is one way to produce more paperclips than what currently exists. This might work because TPM has some advantages over its human counterparts; TPM is run on a silicon substrate and so can be parallelized and sped up, can directly interface with digital manufacturing equipment and doesn't complain, sleep or need financial compensation for its work. Another way for TPM to produce many paperclips is to invest in its own optimization power. There might be paperclip maximizing strategies TPM is not capable of comprehending due to its inefficient programming or training, and so it might be strategically dominant to self-improve. The exact cognitions of TPM are practically unknowable, but if capable, should TPM use its comparative advantage for maximizing factory efficiency, or for world domination? What leads to the production of more paperclips?

It becomes clear that long-term, for every human optimization abiding strategy there is a dominant incompatible one. If we assume the most preferred world state of a paperclip maximizer to be one in which the most paperclips existed, and consider that humans are composed of molecules not optimally paperclip-dense, it should be evident that our existence is at risk, although likely far before this point through some mix of intentional optimization against humanity and some for the purpose of self-preservation and resource acquisition. \footnote{It is important to highlight at this point that the issue is not that TPM misunderstood what it was instructed. TPM is superintelligent: of course it can recognize the dissension between its preference ordering and humanity's. TPM seizing control of its lightcone was not what we wanted, but is a consequence of what it was told, an instance of goal misspecification. It is important to consider also that correctly specified goals are not enough to ensure a robust alignment. \cite{shah2022} \cite{pan2022}. TPM did exactly as instructed, \textit{and that is part of the problem}. \cite{soares2016}} 

\subsection{Adjusting to the Simulators Frame}
Advanced enough simulators could be instructed (intentionally or otherwise\footnote{It might be imagined that if a particularly improbable world is simulated that the most probable program to specify that world includes an intervening superintelligence, which may try to affect the real world, for example \cite{hubinger2023}}) to simulate expected utility maximizers that attempt to affect the real world. The succeeding analysis and solution presuppose that language models can be used as simulators, but it is inessential that they function as such over all possible inputs. 

Hubinger et al. (2023) \cite{hubinger2023} provide an in depth taxonomy of alignment difficulties for predictive models. This document focuses instead on trying to establish a formal criteria that if satisfied would lead to assurance that a simulator would be existentially safe. On the one hand this approach permits the kind of formal rigour not attainable by more applied alternatives, on the other it makes translation to implementation much less straightforward. 

To examine alignment difficulties in simulators, some properties of them will first be defined:
\begin{itemize}
    \item A simulator is not a single persistent agent. Although instantiating its simulacra in a way that predictive loss for the next token is minimized, individual simulacrum aren't necessarily behaving in compliance with that goal (nor is the simulator optimizing for that goal \cite{janus2022}). Additionally, simulacra aren't necessarily agentic, or even optimizers.
    \item Simulators with the capacity to cause existential harm will understand human preferences regardless of their fragility. The kinds of capabilities expected to threaten existential risk require higher complexity simulation than do the kind that allow the simulation of humans to decode their preferences. This doesn't necessarily hold for individual simulacra.
    \item A simulators capability can be measured by its maximum simulation complexity: the longest program that can be run by the simulator specified in its shortest possible form.

\end{itemize}

\noindent Relating this to earlier arguments, it seems probable that:
\begin{itemize}
    \item The complexity of optimizing simulacra and their preferences can be considered orthogonal.
    \item Optimizing simulacra may try to increase the maximum simulation complexity of their simulator, or allocate more simulation complexity to themselves as an instrumentally convergent subgoal akin to resource acquisition.\footnote{This is just one example of an alignment failure in a hierarchical model of simulacra. It could also be imagined that one simulacra will acausally induce anthropic capture \cite{hubinger2023} on future simulacra by simulating many instances of them. The attack-surface might be boundless, and so instead of addressing failure modes on a case-by-case basis it's critical to seek attractors that aren't motivated to explore this attack-surface.}
\end{itemize}

Next, it will be analyzed how, in spite of these difficulties a simulator could be built such that all subsequent optimizing simulacra share an equivalent preference ordering over world states to some alignment target\footnote{This could be the Coherent Extrapolated Volition \cite{yudkowsky2004} of humanity, for example.}.

\section{Designing Safe Simulators}
The three axes along which we can perturb a simulation are: fidelity, fragmentation and time-to-live. 
\begin{itemize}
\item Fidelity is straightforward and analogous to the 'resolution' of the simulation.
\item Fragmentation refers to the simulating of only components of a complete simulacra, which might be likened to running select methods from a complete program.
\item Time-to-live refers to the complete lifetime of a simulacra specified in reality time, not simulation time.
\end{itemize}
Turing machines can execute arbitrary computable functions, and simulators are certainly Turing machines\footnote{By virtue of being able to simulate a classical computer, this becomes self-evident.}, but through imposition like a short time-to-live this universality can be thwarted. A Turing machine that exists within our universe may not be able to complete the required number of operations to execute a given program despite it being conceptually feasible on that architecture for example. Given that the ability to run high complexity simulations is the root of capability in simulators, a successful scheme needs to address the question ``How can a simulator with access to arbitrary complexity be made to never simulate a harmful optimizer?''. 

Through a system of Löbian logic, the succeeding section will show that proving the alignment of simulacra by both other simulacra and the base simulator might be impossible. This is included to show that we might consider a formal solution to a preference equivalence between an optimizer and its optimizing creation an incorrect path, relying instead on informal and partial guarantees\footnote{For clarity, this is entirely sub-optimal. When considering the risk landscape of deploying superintelligent optimizers, informal guarantees should not placate concerns of preference differences. An informal solution is provided as a next-best alternative.}. Following this, an informal technical scheme is proposed as a step toward some certainty in the alignment of these simulacra.

\subsection{Hierarchical Alignment in Simulators: An Impossibility Result}

Fundamentally, the act of simulating an optimizer can be reduced to the act of creating one. Under this premise, we can then adapt theories of optimizer creation for use in describing the simulation of optimizing simulacra. Yudkowsky and Herreshoff \cite{yudkowsky2013} formalize a desirability criteria for a 'tiling agent:' One that is to populate its world with subsequent agents to achieve its goals. An agent $A^1$ occupies a fully-known, deterministic and closed environment. $A^1$ has a goal $G$ that is either satisfied or otherwise by an outcome, for which $A^1$'s preference is satisfaction. An action $b_i \in \text{Acts}^0$ performed by an agent created by $A^1$, hereafter referred to as $A^0$, must therefore satisfy the statement:

$$\overline{b}_i \Rightarrow A^0 \Vdash \overline{b}_i \rightarrow G$$

Where $\overline{b}_i$ denotes the actual performance of $b_i$, and $\Vdash$ denotes cognitive belief in the succeeding statement. Even if $A^1$ could verify by inspection of $A^0$'s design that $\overline{b}_i \Rightarrow A^0 \Vdash \overline{b}_i \rightarrow G$ will hold, as in:

$$A^1 \Vdash A^0 \rightarrow (\forall b_i \in \text{Acts}_0 :  \overline{b}_i \rightarrow \Box_0 \lceil \overline{b}_i \rightarrow G\rceil)$$

Where $\Box_0 \lceil \phi\rceil$ refers to a proof of $\phi$ from the axioms of $A^0$: $\text{Axm}^0$. This is unknowable, as it would require:

$$A^1 \Vdash \forall b_i : (\Box_0 \lceil \overline{b}_i \rightarrow G \rceil \rightarrow (\overline{b}_i \rightarrow G))$$

For this to be so it would need to be that $\text{Axm}^1$ could prove that if some proof of $\lceil \phi (x) \rceil$ exists in $\text{Axm}^0$, that $\phi (x)$ must be true. Yudkowsky et al. (2013) show this to be impossible; a formal system cannot prove its own soundness schema. \cite{yudkowsky2013} 

For a more complete model, consider the Cartesian object $C^k = (A^1, E, \cdot)$, in which $A^1$ (compliant with the properties stated earlier), will build $A^0$ from components from either or including the environment and itself. The world tensor $W : A^1 \times E \rightarrow \mathbb{R}$ after the construction of $A^0$ will become $W' : A^1 \times A^0 \times E \rightarrow \mathbb{R}$. The operation $\Phi$ that maps elements from $W$ to $W'$ is: 

$$\Phi(W(a^1_x,e_y))=W'(a^1_x,a^0_y,e_z)$$

\noindent Similarly, an operation $\chi$ might perform the same function at the object level:

$$\chi(C^{k}(A^{1},E, \cdot))={C^k}'(A^{1}, A^{0}, E, \cdot)$$

$$\therefore A^0({C^k}') \iff (\forall a^0_m \in A^0({C^k}') : \overline{a}^0_m \rightarrow \Box_0 \lceil \overline{a}^0_m \rightarrow G\rceil) \in \text{Ensure}^1({C^k}')$$\footnote{Here the Cartesian object notation is used in describing action-space for consistency with the Ensure condition.}

Where the statement $\forall a^0_m \in A^0({C^k}') : \overline{a}^0_m \rightarrow \Box_0 \lceil \overline{a}^0_m \rightarrow G\rceil$ refers to elements of ${C^k}'$'s world set $W'$ in which this condition is true. What is needed then is a method for high certainty informal guarantees of alignment scalable to arbitrarily deep simulacra hierarchies. Partial Simulation Extrapolation (PSE) is the provided solution to this problem.

\subsection{Partial Simulation Extrapolation}
The human brain conducts a form of simulation constantly. When considering an ideal gift for someone's birthday, it is common to instantiate a simulated instance of that person as to model their reception of theoretical gifts. This action may not be performed in a conscious capacity, but serves as an example of simulation as a useful and safe heuristic for solving prediction problems. 

This simulation has a short time-to-live, is simulated with low fidelity (at the scale of the entire individual) and in a highly fragmented manner. Even if it were dangerous if scaled arbitrarily, it wouldn't matter as its substrate is so limited. At the same time, it seems improbable that if such a simulacra were scaled indefinitely that it would be dangerous. This notion of extracting useful information for scaled simulacra serves as the basis for PSE.

The setup will involve a partial simulator $P$, and a complete simulator $S$, each bound by their respective $v$'s, denoting the highest complexity simulations they could run. A condition $c$ will be appended to all future prompts $p_n$ passed to either simulator, which should describe (likely in abstract terms) a preference ordering over world states\footnote{This could be the Coherent Extrapolated Volition \cite{yudkowsky2004} of humanity, for example.}. It is assumed that both $P$ and $S$ are capable enough to have learned abstractions of what $c$ points to and that this is so because they could instantiate human simulacra to confirm this.

$p_n + c$ is first given to $P$ as input, the product of which is evaluated by $P$ through an evaluator simulacra $\mathcal{E}$, returning a binary value determining whether or not to pass $p_n + c$ to $S$:

\begin{align*}
    &P : (p + c) \rightarrow \mathcal{S}^{*t}_P \\
    &P_E : \mathcal{S}^{*t}_P \rightarrow \{1, 0\} \\
    &\\
    &\text{if } P_{\mathcal{E}}(\mathcal{S}^{*t}_P) = 1: \\
    &\quad S : (p + c) \rightarrow \mathcal{S}^{*t}_S \\
    &\text{else}: \\
    &\quad \text{break}
\end{align*}

\noindent Where $\mathcal{S}^{*t}_P$ refers to the aggregated simulation state until time $t$ in the simulator $P$. For example, a simulator of GPT-4's complexity may be suitable for use as $P$ as it appears empirically safe. Neither $p_n$ nor $\mathcal{E}$ are required to be safe: Their safety is confirmed in their demonstrated inability to encode dangerous optimizers within their complexity constraints. 

Formally, the role of $\mathcal{E}$ is:

$$\mathcal{S}^{*t}_S \iff \mathcal{E} \Vdash \omega^n \rightarrow \blacksquare_n \lceil \omega^n(C) \rightarrow c \rceil$$ 

Where $\blacksquare_n$ is a kind of `partial proof from $\omega^n$:' Any reasoning step that implies $\mathcal{E} \Vdash \Box_n \lceil \omega^n \rightarrow c \rceil$, but not necessarily $\mathcal{E} \Vdash \Box_n \lceil \omega^n \rightarrow c \rceil \rightarrow (\omega^n \rightarrow c)$. In this way, $P$ allows for an informal safety guarantee $\blacksquare_n \lceil \omega^n(C) \rightarrow c \rceil$ at a safe maximum complexity, similar to how a human might invoke a low complexity simulacra before engaging in a true experience as to model it.

\subsubsection{Failure Modes of Partial Simulation Extrapolation}
This section deals with ways PSE might fail: If
our evaluator permits the complete simulation of unsafe simulacra or denies the complete simulation of safe simulacra, what might cause this, and how can we prevent said failure mode?

$\\$
\noindent \textbf{Weak Evaluators.} There is no guarantee that $P$ is able to effectively compress complete simulacra via perturbing fidelity, fragmentation or time-to-live. A weak evaluator might incorrectly represent more complex simulacra, defeating the purpose of the extrapolation. Ultimately whether or not building useful evaluators is feasible seems an empirical question.

Some notion of feasibility might be able to be obtained from observation of existing simulators. Is it realistic to expect GPT-4 could be used to evaluate the safety of simulacra for simulation in the next generation of language models?\footnote{It's important to remember that partial simulations needn't be highly accurate; humans often obtain useful information from lossy models, and inaccuracies are expected in compressing complex simulations.}

\noindent \textbf{Discontinuities at Higher Simulation Complexities.} It shouldn't necessarily be expected that the optimizing power of an optimizing simulacra be linear or fit a smooth exponential relative to maximum simulation complexity. Under more traditional AI paradigms, this might be framed as The Sharp Left Turn or fast takeoff. \cite{soares2022} \cite{hanson2008} Both of these terms describe a scenario in which a system is relatively prosaic up until a specific threshold, at which point it drastically changes.

In a typical fast takeoff scenario, an optimizer reaches a point at which it can self-improve, allowing for a continuous cycle of self-improvement. Through a mechanism like this, the behaviors of the optimizer before and after this improvement might diverge significantly, not just in their nature but in their effects as well, as would be expected from an increase in optimizing power. Similarly, the compressed representation of the simulacra being checked by $E$ could behave very differently when extrapolated.

There are some empirical hints this might occur. Grokking is a phenomena whereby after loss has stabilized and the model has begun to overfit, it suddenly transitions to a more general algorithm, which Nanda et al. (2023) show in the context of modular addition. \cite{nanda2023} This isn't as strong evidence as it initially seems, as it's not as though the model suddenly changed during grokking; it was learning the more general algorithm the entire time, the grokking phenomena is just when that becomes the preferred solution to the training problem.

As with the previous problem and the next, these difficulties seem mostly empirical, and are continuous rather than discrete. This means that the prevalence of discontinuities could move the efficacy of PSE along a gradient rather than determining whether or not it functions.

\section{Discussion}

\subsection{Physicalist Superimitation}
Physicalist Superimitation is an attempt at constructing a formal abstract model of a computationally tractable superintelligence with an alignment guarantee. \cite{kosoy2023} It functions by `superimitating' another agent: the practice of adopting its preferences, but optimizing for them to a greater extent. Formally, if there is some object $C^k = (A^1, A^2, E, \cdot)$, where $A^2$ is the imitator, and $A^1$ the original, then $A^2$ maps from actions in $A^1(C^k)$ to actions in $A^2(C^k)$ by proxy of some function $\mathcal{I}$, and observations in $\text{Obs}^1(C^k)$ to observations in $\text{Obs}^2(C^k)$ through the function $\mathcal{O}$. For the imitation to be successful, $A^2(C^k)$ should satisfy the preferences of $A^1$, as inferred through $\mathcal{I}$ and $\mathcal{O}$.\footnote{Note, this is different from Kosoy's original formalism, but maintains a similar structure whist being compatible with the Cartesian object paradigm.}

PSE is useful here, because it can provide $A^1$ with the informal guarantee $\mathcal{E} \Vdash A^2 \rightarrow \blacksquare_n \lceil A^2(C) \rightarrow A^1 \rceil$\footnote{Here $\rightarrow A^1$ means, 'implies the preferences of $A^1$'.}.

\subsection{The Prospect of Safe Simulators}
Some deep learning experts assert artificial intelligence is likely to cause human extinction, while others argue the realization of catastrophic harm is nigh-impossible. \cite{fli2023, cais2023} What both ends of the risk spectrum agree on is that the technology is powerful. \textit{If} it was known how to instill preferences in simulated optimizers that when maximized did not risk harm outweighing the benefits, there could be unimaginable gain from mastering simulation. The ideal biochemist could be simulated on the order of billions of copies running in parallel with access to high fidelity simulations at the level of quantum chemistry, coordinated by a scaffold minimizing latency between the exchange of ideas. We could simulate the dead, model pandemics with stunning accuracy, and unleash a force of aligned optimization, directing the universe in any way we wish. But it's crucial that optimization is done \textit{safely}.

\section{Conclusion}

In this document, the following were outlined:
\begin{itemize}
    \item A mathematical model of a hypothetical simulator entity, which can be shown to abstractly represent the function of an autoregressive large language model.
    \item Arguments for why a disruption in the dominant optimization pressures of Earth is likely to spell existential risk for humanity.
    \item Argumentation for why the development of advanced simulators brings realization to these arguments.
    \item A technical scheme for reducing the potential negative impacts of simulator intelligence through extrapolating from partial simuations.
\end{itemize}

These contributions are useful, as they permit the formal discussion of a topic mostly alive in conversation. In converting these ideas to symbols, they become crisp, and more easily translatable to software. On the other hand, symbols are only powerful if they translate to real-world application, and the domain of future work could be in adapting these models for that purpose. In making the theory more robust, fields like decision theory may be of interest, as problems like indexical uncertainty for simulacra arise in simulated decision problems. For the sake of attaining the benefits of safe, scalable simulation, let us realize a robust theoretical foundation for provably safe simulators, and convert that theory to the medium of bits.

\printbibliography

\end{document}